\documentclass[sigconf,natbib=false,anonymous=False]{acmart}

\AtBeginDocument{%
 }

\setcopyright{none}
\acmConference{}  
\acmYear{}
\acmISBN{}
\acmDOI{}

\usepackage{algorithmic}
\usepackage{graphicx}
\usepackage{comment}
\usepackage{placeins} 
\usepackage{float} 

\usepackage{tabularx}
\usepackage{rotating}
\usepackage{textcomp}
\usepackage{xcolor}
\usepackage{booktabs}
\usepackage{amsmath} 
\usepackage{amsfonts} 

\usepackage{algorithm}
\usepackage{algpseudocode}
\DeclareMathOperator*{\argmax}{arg\,max}
\DeclareMathOperator*{\argmin}{arg\,min}

\usepackage[
  datamodel=acmdatamodel,
  style=acmnumeric, 
]{biblatex}

\AtEveryBibitem{\clearfield{note}}

\begin{document}

\title{A Matter of Time: Revealing the Structure of Time in Vision-Language Models}

\author{Nidham Tekaya}
\authornote{Corresponding author.}
\orcid{https://orcid.org/0009-0003-8679-4082}
\email{ntekaya@fhstp.ac.at}
\affiliation{%
  \institution{St. Pölten University of Applied Sciences}
  \city{St. Pölten}
  \country{Austria}
}
\affiliation{%
  \institution{TU Wien}
  \city{Vienna}
  \country{Austria}
}

\author{Manuela Waldner}
\orcid{https://orcid.org/0000-0003-1387-5132}
\email{manuela.waldner@tuwien.ac.at}
\affiliation{%
  \institution{TU Wien}
  \city{Vienna}
  \country{Austria}
}

\author{Matthias Zeppelzauer}
\orcid{https://orcid.org/0000-0003-0413-4746}
\email{mzeppelzauer@fhstp.ac.at}
\affiliation{%
  \institution{St. Pölten University of Applied Sciences}
  \city{St. Pölten}
  \country{Austria}
}

\renewcommand{\shortauthors}{Nidham Tekaya, Manuela Waldner, \& Matthias Zeppelzauer}

\begin{abstract}

Large-scale vision-language models (VLMs) such as CLIP have gained popularity for their generalizable and expressive multimodal representations. By leveraging large-scale training data with diverse textual metadata, VLMs acquire open-vocabulary capabilities, solving tasks beyond their training scope. This paper investigates the temporal awareness of VLMs, assessing their ability to position visual content in time. We introduce TIME10k, a benchmark dataset of over 10,000 images with temporal ground truth, and evaluate the time-awareness of 37 VLMs by a novel methodology. Our investigation reveals that temporal information is structured along a low-dimensional, non-linear manifold in the VLM embedding space. Based on this insight, we propose methods to derive an explicit ``timeline'' representation from the embedding space. These representations model time and its chronological progression and thereby facilitate temporal reasoning tasks. Our timeline approaches achieve competitive to superior accuracy compared to a prompt-based baseline while being computationally efficient. All code and data are available at \url{https://tekayanidham.github.io/timeline-page/}.


\vspace{0.5em}
\noindent\textit{Note:} © ACM 2025. This is the author's version of the work. It is posted here for your personal use. Not for redistribution. The definitive Version of Record was published in Proceedings of the 33rd ACM International Conference on Multimedia (MM '25), \url{http://dx.doi.org/10.1145/3746027.3758163}.

\end{abstract}



\begin{teaserfigure}
    \centering
    \includegraphics[width=\textwidth]{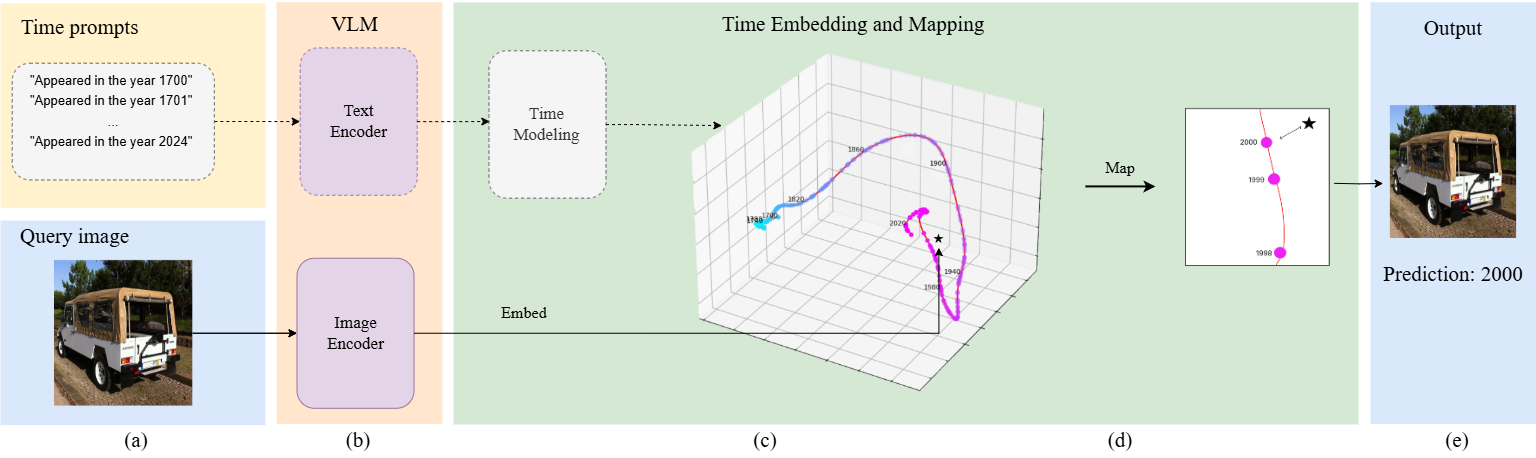}
    \caption{Overview of our approach to time assessment in images. (a) Time-specific prompts and query images are input to the VLM. (b) Text and image encoders generate embeddings in the shared multimodal space. (c) Time embeddings form a chronological manifold that we model as a 1D curve in a subspace of the embedding space. (d) Image embeddings are mapped to the closest year on the time curve. (e) Time predictions (in years) are output for each query image.
    }
    \label{fig:teaser}
\end{teaserfigure}


\maketitle

\section{Introduction}
\label{sec:intro}

Time plays a fundamental role in the analysis of historical image and video collections and represents an essential aspect in cultural studies, historical analysis, and the management of digital archives. Determining when certain artifacts first appeared in time enables answering questions, such as: \textit{how old is this object?} and \textit{when did it first appear?} It further provides valuable constraints for dating historical images, i.e. \textit{what is the earliest possible time this picture was taken?}. From a technical perspective, it is important to examine whether vision-language models (VLMs) exhibit inherent temporal reasoning. If so, understanding how this reasoning is encoded in their learned representations is key to evaluating their ability to infer the temporal origin of objects.


Despite advances in visual analysis and understanding, dedicated methods for time assessment remain underdeveloped due to time's abstract nature and a lack of datasets with sufficient and adequate metadata. We investigate whether VLMs implicitly capture temporal structures and if they provide a certain awareness to time. To the best of our knowledge, this is the first analysis of this kind.

With the emergence of VLMs being trained on large-scale image datasets \cite{christoph_schuhmann_laion-400m_2021, schuhmann_christoph_laion-5b_2022} and paired with textual metadata, powerful representations have become available. These pre-trained VLMs can map concepts to images and vice versa, facilitating numerous downstream tasks \cite{xu2023cit}. The metadata used for training VLMs is often derived from web sources, including associated text such as alternative image descriptions from HTML or other contextual annotations \cite{schuhmann_christoph_laion-5b_2022, srinivasan2021wit, fang2023data}, which may implicitly encode absolute or relative temporal information like object descriptions (e.g., ``a vintage car from the 1960s'' or ``an early model of a mobile phone''), event contexts (e.g., ``photographed during the 1980 Olympic Games''), or references to time periods (e.g., "artwork from the Renaissance period"). While the exact nature and quality of these annotations can vary significantly, we can assume that the learned representations implicitly capture a certain amount of temporal information and context, making the models to some degree time-aware.

This paper is the first to systematically explore the time-awareness and temporal modeling capabilities of VLMs. By analysing their embedding spaces, we investigate how these models implicitly encode temporal information. Our investigation specifically focuses on \textit{human-made artifacts} and on the ``\textit{time of first appearance}'' defined as the time point when a given object was first introduced in history or became publicly available (e.g., the year when a new car model was released). Note that, our goal is not to achieve perfect accuracy on exact release dates, but more importantly to assess whether models can approximate these temporal landmarks within a reasonable margin, providing meaningful temporal context. In our investigation, we particularly focus on contrastively trained VLMs such as CLIP and its successors~\cite{li2023blip, liu2023visual, chen2023pali} with open-vocabulary capabilities. These architectures represent the backbones for most recent foundation VLMs, including also generative models like~\cite{vardi2025clip, lerner2024cross, lin2024clip}. To cover a broad spectrum of architectures, we include ImageBind \cite{rohit_girdhar_imagebind_2023}, ViT-Lens \cite{weixian_lei_vit-lens_2023}, EVA-CLIP \cite{qiang_sun_eva-clip_2023, quan_sun_eva-clip-18b_2024}, and SigLIP \cite{xiaohua_zhai_sigmoid_2023}.

Our investigation is guided by the following central research questions:

\begin{itemize}
    \item RQ1: To what extent are open-vocabulary VLMs inherently time-aware?
    
    \item RQ2: How is the temporal information of images and text structured in the embedding space?
    
    \item RQ3: If temporal structure exists (RQ2), how can we represent time to effectively and efficiently predict the time of first appearance of the images?
\end{itemize}

There are a number of challenges to tackle to answer our research questions: (i) there is limited availability of datasets with explicit time annotations which makes the systematic comparison and performance benchmarking difficult; (ii) VLM embedding spaces are high-dimensional (typically 512+ dimensions) with latent dimensions that lack semantic interpretation, making it challenging to identify which dimensions encode temporal information; (iii) prompting VLMs for temporal context (i.e. prompt probing~\cite{lewis2022does}) is possible (due to their open vocabulary nature), however, it requires generating embeddings for each year, limiting its scalability for larger temporal ranges.

 To address the above challenges, this paper introduces a novel methodology for systematically exploring and leveraging the temporal reasoning capabilities of VLMs. Our main contributions are as follows:
\begin{itemize}
    \item We introduce \textbf{TIME10k}, a temporally annotated dataset with over 10,000 images from 6 classes of objects, enabling systematic evaluation and comparison of VLMs with respect to temporal awareness and time prediction capabilities.
    \item We propose a comprehensive framework for objectively evaluating time-awareness and investigate 37 state-of-the-art VLMs, examining various backbones, architectures, and prompting strategies, and providing a clear pathway for benchmarking temporal reasoning.

    \item A novel approach for deriving an explicit model of time from a VLM's embedding space in the form of a sequential ``timeline'' representation, as illustrated in Figure~\ref{fig:teaser}, which can be leveraged to effectively and efficiently place images in a temporal context.

\end{itemize}

\section{Related Work}

Related work for our work comes from different areas, briefly reviewed in the following.

\textbf{Open-Vocabulary Vision-Language Models.} VLMs have gained traction since the introduction of CLIP~\cite{alec_radford_learning_2021}, which uses contrastive training to align image-text pairs and enables open-vocabulary learning. The result is a shared latent space where semantically related images and text are closely positioned in terms of dot-product similarity. CLIP has inspired the development of models such as OpenCLIP~\cite{mehdi_cherti_reproducible_2023}, EVA-CLIP~\cite{qiang_sun_eva-clip_2023, quan_sun_eva-clip-18b_2024}, SIGLIP~\cite{xiaohua_zhai_sigmoid_2023}, ImageBind~\cite{rohit_girdhar_imagebind_2023}, and ViT-Lens~\cite{weixian_lei_vit-lens_2023}, which integrate textual and visual data and even additional modalities. These models demonstrate strong zero-shot classification~\cite{qizhou_wang_sober_2024} and cross-modal retrieval capabilities~\cite{cherag_aroraa_smart_2024}. Researchers have examined how CLIP embeddings encode perceptual attributes~\cite{wang_jianyi_exploring_2022} and geometric reasoning~\cite{danier2024depthcues}. Building upon these lines of investigation, our work analyses whether VLMs implicitly encode temporal information about depicted objects.

\textbf{Textual Modeling of Temporal Aspects.}
Recent work has demonstrated that language models learn structured temporal representations from textual data. Gurnee \& Tegmark~\cite{heinzerling2024monotonic} identified ``time neurons'' that encode temporal coordinates in linear representations when processing text about historical figures, while Heinzerling \& Inui~\cite{gurnee2023language} found that temporal properties like birth years are encoded in low-dimensional linear subspaces and demonstrated control by editing internal model activations to systematically change the model's temporal outputs. Although highly related to our work,  these approaches focus only on the textual modality and LLMs. Our work investigates temporal context across the visual and the textual domain in VLMs. Interestingly, while the above works find time to be modelled linearly, our work reveals a \emph{non-linear} structure of time in VLMs.

\textbf{Visual Modeling of Temporal Aspects.}
Temporal aspects in images have been studied for face aging~\cite{ziyu_wan_bringing_2020}, historical portrait generation~\cite{xiao_dong_unidiff_2023}, and video modeling~\cite{xi_ding_language_2024}. Chen et al.~\cite{eric_ming_chen_whats_2023} used StyleGAN to synthesize faces across historical periods (1880s-2020s) through stylistic transformations. Methods have leveraged explicit temporal metadata from LAION-5B~\cite{schuhmann_christoph_laion-5b_2022, li2024datacomp, chen2022pali,fang2023data} for historical retrieval. Most related is Barancová et al.~\cite{alexandra_barancova_blind_2023}, who explored OpenCLIP for zero-shot dating of historical photographs from 1950-1999~\cite{artidigh21}, showing bias towards earlier years and improved accuracy via fine-tuned logistic regression. We go beyond this study and investigate \textit{intrinsic} temporal properties across 37 VLMs without fine-tuning and introduce methods for deriving an \textit{explicit} timeline representation from embedding spaces.

\textbf{Embedding Space Analysis.} Understanding high-dimensional embeddings is crucial for the interpretability of the models. To improve it, Concept Activation Vectors (CAVs) \cite{a_nicolson_explaining_2024} define semantic directions in embedding spaces, while PCA-based GANSpace \cite{erik_harkonen_ganspace_2020} uncovers latent structures related to semantic attributes such as pose, lighting, ageing, and object transformations in generative models. Furthermore, dimensionality reduction techniques such as UMAP~\cite{leland_mcinnes_umap_2018} have been applied to explore contrastive embeddings, for example in~\cite{jiazhi_xia_interactive_2022} for interactive cluster analysis in contrastive spaces. However, to the best of our knowledge, prior work has not examined whether temporal information manifests in some kind of structure in embedding spaces.

\section{Methodology}

In the following, we outline our methodology.  
We first describe \textit{Time Probing}, a baseline for temporal inference (Section~\ref{sec:prompt_probing}). Next, we analyse the spatial structure of temporal embeddings in the VLM's embedding space to shed light on the time-awareness of VLMs (Section~\ref{sec:embedding_analysis}) and finally we propose a method to derive an explicit representation of time from the embedding space that enables efficient temporal inference as an alternative to time probing (Section~\ref{sec:timeline_representation}). Note that in the following we refer to \textit{time embeddings} as the outputs of the VLM's text encoder when prompted with time-specific queries, which represent specific years and serve as temporal anchors within the shared latent space.

\subsection{Time Probing: A Baseline for Retrieving Temporal Context (RQ1, RQ2)\label{sec:prompt_probing}}

\begin{figure}[ht]
    \centering
    \includegraphics[width=0.95\columnwidth]{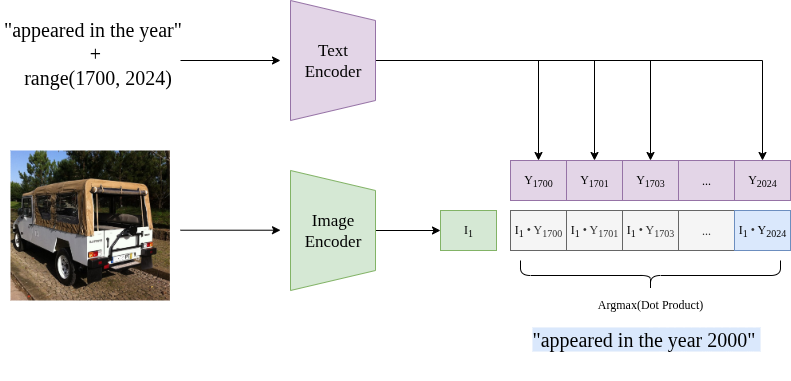} 
    \caption{Time probing: temporal inference via dot-product similarity. Time embeddings (here for years 1700-2024) are compared with image embeddings. The time embedding with the highest similarity indicates the year of first appearance.}
    \label{fig:dot_product_inference}
\end{figure}

Time probing is a prompt-based baseline inspired by Barancová et al.~\cite{alexandra_barancova_blind_2023}, who used VLMs for zero-shot dating of photographs. We adapt this approach to estimate when depicted objects first appeared by exploiting the cross-modal alignment in the VLM embedding space (Figure~\ref{fig:dot_product_inference}).
 Time probing generates text prompts by combining a template sentence with a year $y \in Y$ for a given time range $Y = [Y_{\text{min}}, \ldots, Y_{\text{max}}]$. To refer to the ``time of first appearance'', prompts may be defined e.g. as ``First appeared in the year $y$'' or ``Was introduced in the year $y$''. These prompts are encoded using the VLM's text encoder, resulting in \textit{time embeddings} $\mathbf{T}_y$ for individual years. Similarly, the query image is encoded into an \textit{image embedding} $\mathbf{I}$ using the image encoder. Both embeddings exist in the same shared latent space, enabling direct comparison via dot-product similarity. The predicted year $y_{\text{pred}}$ is finally derived as the year whose time embedding has the highest similarity:
\begin{equation}
\label{eq:probe_pred}
y_{\text{pred}} = \argmax_{y \in Y}(\mathbf{I}^{\mathrm{T}} \cdot \mathbf{T}_y),
\end{equation}
where $\mathbf{I} \in \mathbf{R}^N$ is the high-dimensional image embedding, $\mathbf{T}_y \in \mathbf{R}^N$ is the time embedding for year $y$, and $Y$ is the set of candidate years.
Time probing serves as a baseline in our experiments, particularly for answering RQ1. It leverages the cross-modal matching capabilities of VLMs to assess temporal information implicitly captured during training. However, time probing is limited in that it only performs local analysis, treating each year independently without modeling temporal progression or dependencies between time points. It further scales poorly with the number of years in $Y$.
\subsection{Embedding Space Analysis (RQ2)\label{sec:embedding_analysis}}

Time probing treats each year's embedding independently, ignoring potential relationships between time points. This prevents us from capturing higher-level temporal structure. To advance beyond simple probing, we need to understand how time is organised within the VLM's embedding space, which is the central focus of RQ2. The spatial distribution of the time embeddings in the embedding space is a priori unknown. Time points could follow linear or non-linear relationships, form clusters, or lack any coherent structure. To investigate the spatial structure of time embeddings we leverage dimensionality reduction techniques. Thereby, we proceed in two ways: First, we visualize lower-dimensional projections of time embeddings to identify potential structures. Second, we generate one-dimensional projections to measure how well individual years align chronologically, to test whether a temporal order exists among the embeddings. For our analysis, we select the following dimensionality reduction techniques:

\noindent
\textbf{Kernel PCA:} (KPCA) \cite{alberto_garcia-gonzalez_kernel_2020} is chosen for two reasons: First, it enables the use of the cosine kernel, which directly aligns with the metric used in contrastively trained VLMs and thereby ensures consistency with the intrinsic structure of the embedding space. Second, it preserves the global structure of the underlying space and does not introduce local distortions, thereby avoiding overfitting. 

\noindent
\textbf{UMAP:} \cite{leland_mcinnes_umap_2018} is a topological dimensionality reduction approach that aims to preserve most of the global structure of the space but not necessarily the local metric structure. It is thus complementary to KPCA. UMAP actually preserves the topology and tries to discover a manifold on which the input data are uniformly distributed~\cite{mcinnes2018umap}. This makes UMAP a promising approach for testing whether temporal information forms a structured manifold in the embedding space. Additionally, UMAP's ability to transform and embed new data points into a previously computed projection (crucial for projecting image embeddings) makes it a better choice to alternatives like t-SNE~\cite{laurens_van_der_maaten_visualizing_2008} for our application.

We generate 1D, 2D, and 3D projections of time embeddings $\mathbf{T}_{y_{\min},...,y_{\max}}$ to analyse temporal structures and assess the chronological ordering by comparing the sequence of the 1D projections with the respective ground truth sequence.

To further investigate the \textit{alignment} of temporal information across modalities (RQ2), we project image embeddings (of time-annotated images) into the low-dimensional subspaces that we previously generated for the time embeddings with the above dimensionality reduction techniques. If temporal information is aligned across the modalities, projections of image and time embeddings should be close in space.

\subsection{Chronological Timeline Modelling (RQ3)\label{sec:timeline_representation}}
\vspace{-2pt}

Time probing (see Section~\ref{sec:prompt_probing}) treats each year independently, ignoring chronological relationships between temporal embeddings and failing to exploit temporal structure. The intuition behind our proposed approaches is that if temporal information forms a low-dimensional manifold in the embedding space, as our analysis suggests, we can exploit this structure to create an explicit ``timeline'' representation. Whether this representation maintains coherent chronological ordering is an open question that we aim to answer.

We propose two approaches for deriving a sequential time representation from time embeddings: (i) an implicit approach based on UMAP combined with hyperparameter optimisation and (ii) an explicit approach based on Bézier curve approximation. Both approaches offer alternatives to direct probing, with the Bézier curve \cite{zhao2013parameter} explicitly modeling temporal progression in a smooth and coherent manner and UMAP leveraging its ability to unfold non-linear manifolds \cite{leland_mcinnes_umap_2018} and being guided to organise time embeddings in a chronological order.
Note that we just require the order of time embeddings to be chronological but not their sequence to be linear, i.e., equally spaced. We refer to such time representations in the following as ``timelines''. 

\subsubsection{UMAP-Based Timeline Representation}
\label{sec:umap}

We apply UMAP to project the high-dimensional time embeddings $\mathbf{T}_y, y \in Y$ onto a one-dimensional timeline, to unfold the non-linear manifold of time embeddings. This process consists of three key steps:

\textbf{1. Optimising UMAP parameters:}  
UMAP has a number of hyperparameters (e.g., \textit{n\_neighbors} controlling local neighborhood size, \textit{min\_dist} controlling point separation in the embedding, and \textit{metric} specifying the distance measure used). It is a priori not clear which parameters best unfold the manifold of time embeddings. Thus, we perform parameter optimisation  to learn a transformation \( f_{\text{UMAP}} \) that best maps time embeddings \( \mathbf{T}_y \) to their one-dimensional projections \( \hat{\mathbf{T}}_y \). Thereby, UMAP parameters are optimised using the Tree-structured Parzen Estimator (TPE) method \cite{shuhei_watanabe_tree-structured_2023}, which maximizes Spearman's rank correlation~\cite{wissler1905spearman} between the projected order and the correct chronological order of years.

The transformation \( f_{\text{UMAP}} \) maps high-dimensional time embeddings $\mathbf{T}_y$ to a one-dimensional timeline as follows:
\[
\small
f_{\text{UMAP}}: \mathbf{R}^N \to \mathbf{R}^1, \quad \hat{\mathbf{T}}_y = f_{\text{UMAP}}(\mathbf{T}_y), \quad y \in Y,
\]
where \( \mathbf{T}_y \in \mathbf{R}^N \) represents the time embedding for year \( y \) in the VLM's embedding space of dimension \( N \), and \( \hat{\mathbf{T}}_y \in \mathbf{R}^1 \) is its one-dimensional projection. TPE optimises the UMAP parameters by maximizing the following objective function: $\max \rho(Y, \hat{Y})$, where \( \rho \) denotes Spearman’s rank correlation \cite{wissler1905spearman}, which quantifies the monotonic relationship between the chronological order of years \( Y \) and their projected order \( \hat{Y} \).

\textbf{2. Projecting image embeddings:}  
The learned transformation \( f_{\text{UMAP}} \) is applied to image embeddings \( \mathbf{I} \), yielding their corresponding position on the timeline $\hat{\mathbf{I}}$.

\textbf{3. Temporal inference:}  
To retrieve the most likely year for an image, we identify the closest time embedding to its projection $\hat{\mathbf{I}}$ in the one dimensional projection space using Euclidean distance.

\subsubsection{Bézier Curve-Based Timeline Representation}
\label{sec:bezier}
Alternatively to learning a suitable projection (which is time-consuming), we propose to model time via Bézier curve approximation in the original embedding space or a subspace of it.
The Bézier curve approach provides a structured representation of time that enforces chronological consistency while capturing non-linear temporal relationships. Unlike time probing, which treats each year independently, and UMAP, which learns an implicit mapping, the Bézier curve explicitly models temporal progression, offering a transparent and inherently interpretable approach. 

In a first step, we approximate the chronological sequence of time embeddings ($\mathbf{T}$) with a one-dimensional Bézier curve \( \mathcal{C}(t) \) that is embedded in \(\mathbf{R}^N\), the embedding space of $\mathbf{T}$ (or a projection space \(\mathbf{R}^S\) with dimension \(S < N\) to reduce potential noise in temporal modeling and reduce computational complexity). The curve is fitted to all time embeddings from \( y_{\min} \) to \( y_{\max} \) and is discretized with $N_{samples}$, i.e., $t=1,..,N_{samples}$. We leverage the \textit{de Casteljau algorithm}~\cite{boehm1999casteljau} to interpolate the curve between control points \(\mathcal{P}\) uniformly sampled from the sorted time embeddings. The number of control points \( \mathcal{K} \) trades off over- vs. underfitting. Each embedding \( \mathbf{T}_y \) is projected onto its closest point on \( \mathcal{C}(t) \) by: 
\begin{equation}
\small
\hat{\mathbf{T}}_y = \argmin_{t \in [0,1]} d(\mathbf{T}_y, \mathcal{C}(t)),
\end{equation}
where \(d(.,.)\) denotes Euclidean distance. To align image embeddings with the timeline, they are projected into the same space as the Bézier curve. This can be applied in the original embedding space or in a derived subspace of any dimension $S$ (as long as the projection function $g$ is known, i.e. $\hat{\mathbf{I}} = g(\mathbf{I})$ where \(g: \mathbf{R}^N \to \mathbf{R}^{S}\)). 

For temporal inference, we propose two strategies: (i) nearest-neighbor matching, which assigns the closest projected time embedding, and (ii) interpolation between adjacent time points, i.e.:
\begin{equation}
\small
y_{\text{pred}} = (1 - w) y_{\text{before}} + w y_{\text{after}}, 
\quad w = \frac{d(\hat{\mathbf{I}}, \hat{\mathbf{T}}_{\text{before}})}
{d(\hat{\mathbf{I}}, \hat{\mathbf{T}}_{\text{before}}) + d(\hat{\mathbf{I}}, \hat{\mathbf{T}}_{\text{after}})},
\end{equation}
where \( \hat{\mathbf{T}}_{\text{before}} \) and \( \hat{\mathbf{T}}_{\text{after}} \) are the two closest projected time embeddings before and after \( \hat{\mathbf{I}} \), and \( y_{\text{before}}, y_{\text{after}} \) are their corresponding years. Since interpolation yields fractional year values, this method allows for finer temporal granularity.

\section{Dataset Curation}
\label{sec:dataset}

To evaluate the VLMs awareness to time, we compiled \textbf{TIME10k}, a dataset of over 10,000 time-annotated images showing human-made objects spanning a time range from 1715 to 2024 and six object categories: \textit{Aircraft}, \textit{Cars}, \textit{Instruments}, \textit{Mobile phones}, \textit{Ships}, and \textit{Weapons \& Ammunition}. TIME10k was curated using Wikipedia \cite{wikipedia:homepage} and Wikimedia Commons \cite{wikimedia:homepage},  leveraging Wikipedia’s category system for object introduction dates. We extracted time-tagged objects and retrieved corresponding images, with manual verification to ensure quality. We adopted year-level granularity for temporal annotations as this yields the most consistent temporal categorization of the data and in most cases only years were available. Wikipedia's category system predominantly organises objects by their year of introduction (e.g., "Cars introduced in 2022"), and does not consistently provide finer-grained information. Overall, we consider a granularity of years a good tradeoff for evaluating a VLM's ability to position objects in time.

The dataset contains 10,091 images with varying sample numbers across classes (Cars: 4,393; Mobile Phones: 4,337; Ships: 841; Instruments: 436; Aircraft: 69; Weapons: 15). These object categories were selected based on the availability of time annotations in Wikipedia's hierarchical category system, ensuring reliable temporal annotations. The dataset shows a natural temporal bias toward more recent decades, with pre-1830s images primarily being drawings rather than photographs. This imbalance reflects historical developments and does not limit our evaluation approach.  Furthermore, different object classes cover different time ranges, which can also be considered a real-world bias. 

\section{Experimental Setup}

Here we detail the experimental setup for the evaluation of time-awareness in VLMs and to answer research questions RQ1-RQ3. 
We utilize TIME10k (Section~\ref{sec:dataset}) for all experiments, with images resized and normalised according to the specific VLM specifications. 

\subsection{Prompt Design}
\label{subsec:promtDesign}

To examine the sensitivity of VLMs to linguistic variations in temporal queries, we evaluate multiple prompt formulations (P1-P9) that explore different ways of referencing the first appearance of an object. The prompts are detailed in Table~\ref{tab:prompt_sensitivity}. Additionally, we include minimalistic prompts (P1, P2) to assess the model's ability to infer time only from the indication of a year.

\subsection{Vision Language Models}
\label{subsec:VLMs}
We evaluate a diverse selection of 37 VLMs with different architectures, backbones, and pretraining datasets. The models are applied in a zero-shot manner with frozen backbones (no fine-tuning). A detailed list of the selected VLMs, including the architectures and their chosen backbones is provided in Table~\ref{tab:vlms}.


\subsection{Parameters}
\label{subsec:parameters}

For \textit{time probing} the pre-defined time range $Y$ is \( [Y_{\text{min}}, Y_{\text{max}}] \) where $Y_{\min} = 1700$ and $Y_{\max} = 2024$, resulting in $|Y| = 325$ individual years. No softmax was applied to the similarity scores. The evaluated text prompts are listed in detail in Table~\ref {tab:prompt_sensitivity}, and the evaluated VLMs in Table~\ref{tab:vlms}.


For \textit{embedding space analysis}, UMAP parameters are optimised via TPE \cite{shuhei_watanabe_tree-structured_2023}. The optimisation yields the following parameters for the evaluated models: CLIP (ViT-B/32; $n_{\text{neighbors}} = 38$, $\text{min\_dist} = 0.7446$) and EVA-CLIP (EVA02-CLIP-L-14-336; $n_{\text{neighbors}} = 21$ and $\text{min\_dist} = 0.1040$). We consistently use cosine metric for UMAP.



For \textit{timeline modeling}, we apply UMAP to obtain one-dimensional projections following the same optimisation procedure as in embedding space analysis. The Bézier curve is fitted using \( \mathcal{K} = 200 \) control points sampled uniformly along the time embeddings for the time range $Y$.
We choose a relatively high \( \mathcal{K} \) compared to the overall number of time embeddings of 325 to guarantee fidelity to the potentially non-linear shape of the time embeddings. For curve resolution we set  \( N_{\text{samples}} = 1000 \) to allow for fine-grained interpolation along the curve. 

\subsection{Performance Metrics}
\label{sec:metrics}

We employ two types of metrics: ranking metrics to evaluate the quality of chronological orderings and accuracy metrics that assess time prediction quality. As ranking metrics we employ Spearman's Rank Correlation ($\rho$)~\cite{wissler1905spearman}, Kendall's Tau ($\tau$)~\cite{stepanov2015kendall}, and Modified Normalised Damerau-Levenshtein Distance ($\delta_{\text{MNDL}}$) (adapted from~\cite{miller2009levenshtein}). $\delta_{\text{MNDL}}$ measures ranking errors based on adjacent swaps: $\delta_{\text{MNDL}} = 1 - 2 \frac{S}{M}$, where $S$ is the total number of adjacent swaps required to transform the predicted ranking into the correct order, and $M = \frac{N (N - 1)}{2}$ is the maximum possible swaps for $N$ elements. We adjust this metric to have the same range and meaning as $\rho$ and $\tau$. All metrics range from -1 to 1, where 1 indicates perfect chronological ordering and -1 indicates completely reversed ordering, both considered structurally correct as they maintain temporal relationships. These ranking metrics complement the following accuracy metrics by providing a comprehensive assessment of how well VLMs understand temporal relationships and chronological ordering, beyond just predicting specific years.

For accuracy evaluation, we use Mean Absolute Error (MAE) which measures the average deviation between predicted and ground truth years. Furthermore, we introduce TAI as a time-adaptive accuracy measure that applies greater tolerance for older years (reflecting inherent uncertainty in historical documentation) and stricter tolerance for more recent years. TAI is defined by a tolerance threshold function \(T(y)\)  and an intolerance threshold function \(I(y)\) and performs linear interpolation between them. For a given image, TAI is computed as follows:

\begin{equation}
\text{TAI}_i = 
\begin{cases} 
1, & \text{if } |y_{\text{pred}} - y_{\text{GT}}| \leq T(y), \\
1 - \frac{|y_{\text{pred}} - y_{\text{GT}}|}{I(y)}, & \text{if } T(y) < |y_{\text{pred}} - y_{\text{GT}}| < I(y), \\
0, & \text{if } |y_{\text{pred}} - y_{\text{GT}}| \geq I(y).
\end{cases}
\end{equation}

Both thresholds vary over time. Their start and end values are explicitly defined (in years). In our case we set:
 \(T_{Y_{\text{min}}}=20\), \(I_{Y_{\text{min}}}=50\), \(T_{Y_{\text{max}}}=5\), \(I_{Y_{\text{max}}}=15\).  These setting provide 20-year tolerance for historical versus 5-year for modern objects, which we consider a reasonable tradeoff for our investigation.

\subsection{Evaluation Framework}

Our evaluation consists of three main experiments: time probing, embedding space analysis, and chronological timeline modelling.

\paragraph{Time Probing (RQ1)}
To assess temporal awareness, we perform time probing on 37 VLMs (Table~\ref{tab:vlms}) using different text prompts (see Table~\ref{tab:prompt_sensitivity}). We further analyse the prompt sensitivity (Section~\ref{sec:prompt_probing}) to examine its impact on predictions and scores. 

\paragraph{Embedding Space Analysis (RQ2)}
To investigate the spatial structure of time embeddings, we apply dimensionality reduction techniques (Section~\ref{sec:embedding_analysis}) and evaluate their chronological alignment with the ground truth using ranking metrics. We further examine how image embeddings align with time embeddings to assess cross-modal consistency.

\paragraph{Chronological Timeline Modelling (RQ2 \& RQ3)}
We evaluate timeline modelling using two representative VLMs: ViT-B/32 (CLIP) as the most popular contrastive model and EVA02-CLIP-B/16 (EVA-CLIP) for its strong performance (see Table~\ref{tab:vlms}). This focused selection serves as a proof-of-concept and reduces computational costs. Timeline quality is assessed via ranking metrics for chronological alignment (RQ2) and accuracy metrics (TAI, MAE) for time prediction (RQ3).

\paragraph{UMAP-Based Timeline Representation}
UMAP is optimised to preserve temporal order and is then applied to image embeddings. Predictions are made by mapping images to the closest year. UMAP hyperparameters are detailed in Section~\ref{subsec:parameters}.

\paragraph{Bézier Curve-Based Timeline Representation}
We evaluate Bézier timeline fitting in both high-dimensional and KPCA reduced subspaces. Time inference is performed using nearest neighbour and interpolation (see Section~\ref{sec:bezier}). For visualisation, we reduce embeddings to \(\mathbf{R}^3\) before curve fitting. 

\section{Results}
\label{sec:results}

We first present results for time probing on a broad set of VLMs (Section~\ref{subsec:result_time_probing}). Next, we present our analysis of temporal structures in the embedding space in Section~\ref{subsec:results_embeddinganalysis} and finally, we demonstrate the capabilities of VLMs for timeline modeling (Section~\ref{subsec:result_timeline}).

\subsection{Time Probing}
\label{subsec:result_time_probing}

\textbf{Quantitative results:}
Table~\ref{tab:vlms} presents the MAE and TAI of 37 VLMs using prompt P7 (see Table~\ref{tab:prompt_sensitivity}) on the TIME10k dataset. OpenCLIP (``ViT-bigG-14-quickgelu'') and EVA-CLIP (``EVA02-CLIP-L-14-336'') achieve the best performance, followed by ImageBind \cite{rohit_girdhar_imagebind_2023} and ViT-Lens \cite{weixian_lei_vit-lens_2023}. Overall,
model architecture and training dataset have a strong influence on results. Some models fail at time modeling, possibly due to insufficient training datasets, e.g., ``V1,'' ``CommonPool,'' and ``DFN2B'' underperform, while ``openai'' and ``Merged-2B'' excel. Backbone complexity also matters, e.g., ``ViTamin-S'' performs poorly compared to its larger ``ViTamin-XL-384'' counterpart.

\begin{table}[]
\tiny
\centering
\caption{Evaluation of the awareness of time (in terms of MAE and TAI) across different VLMs with varying network architectures and backbones, pretrained on different data.}
\label{tab:vlms}
\renewcommand{\arraystretch}{1.2}
\resizebox{\columnwidth}{!}{%

\begin{tabular}{lcccc}
\hline

\textbf{Architecture} & \textbf{Backbone} & \textbf{Training Data} & \textbf{$\downarrow$MAE} & \textbf{$\uparrow$ TAI} \\ 
\hline
CLIP \cite{alec_radford_learning_2021}  & RN101 & openai & 10.61 & 0.70 \\
CLIP \cite{alec_radford_learning_2021}  & RN50 & openai & 10.16 & 0.71 \\
CLIP \cite{alec_radford_learning_2021}  & RN50x16 & openai & 9.40 & 0.73 \\
CLIP \cite{alec_radford_learning_2021}  & RN50x4 & openai & 10.61 & 0.72 \\
CLIP \cite{alec_radford_learning_2021} & RN50x64 & openai & 7.26 & 0.81 \\
CLIP \cite{alec_radford_learning_2021}  & ViT-B/16 & openai & 7.75 & 0.79 \\
CLIP \cite{alec_radford_learning_2021}  & ViT-B/32 & openai & 9.53 & 0.70 \\
CLIP \cite{alec_radford_learning_2021}  & ViT-L/14 & openai & 6.99 & 0.83 \\
CLIP \cite{alec_radford_learning_2021}  & ViT-L/14@336px & openai & \textbf{6.76} & \textbf{0.84} \\
\hline
EVA-CLIP \cite{qiang_sun_eva-clip_2023} & EVA01-CLIP-g-14 & LAION-400M & 8.87 & 0.78 \\
EVA-CLIP \cite{qiang_sun_eva-clip_2023} & EVA01-CLIP-g-14-plus & Merged-2B & 11.00 & 0.80 \\
EVA-CLIP \cite{qiang_sun_eva-clip_2023} & EVA02-CLIP-B-16 & Merged-2B & 7.67 & 0.82 \\
EVA-CLIP \cite{qiang_sun_eva-clip_2023} & EVA02-CLIP-L-14 & Merged-2B & 6.30 & 0.86 \\
EVA-CLIP \cite{qiang_sun_eva-clip_2023} & EVA02-CLIP-L-14-336 & Merged-2B & \textbf{6.20} & \textbf{0.86} \\
\hline
EVA-CLIP-18B \cite{quan_sun_eva-clip-18b_2024} & EVA-CLIP-18B & Merged-2B+ & 6.45 & 0.86 \\
EVA-CLIP-18B \cite{quan_sun_eva-clip-18b_2024}  & EVA-CLIP-8B & Merged-2B & 6.53 & 0.84 \\
EVA-CLIP-18B \cite{quan_sun_eva-clip-18b_2024}  & EVA-CLIP-8B-plus & Merged-2B & 6.48 & 0.85 \\
\hline
ImageBind \cite{rohit_girdhar_imagebind_2023} & ViT-H & LAION-2B & 7.16 & 0.83 \\
\hline
CoCa \cite{yu2022coca} & coca-VIT-L-14 \cite{ilharco_gabriel_2021_5143773} & MSCOCO+LAION2B & 8.49 & 0.78 \\
\hline
MobileCLIP \cite{vasu2024mobileclip} & MobileCLIP-S1  & DataComp-DR & 9.70 & 0.75 \\
\hline
ViTamin \cite{jieneng_chen_vitamin_2024} & ViTamin-S \cite{ilharco_gabriel_2021_5143773}  & DataComp-1B & 75.80 & 0.48 \\
ViTamin \cite{jieneng_chen_vitamin_2024} & ViTamin-XL-384 \cite{ilharco_gabriel_2021_5143773}  & DataComp-1B & 6.48 & \textbf{0.86} \\
\hline
OpenCLIP \cite{mehdi_cherti_reproducible_2023}  & RN50-quickgelu \cite{ilharco_gabriel_2021_5143773} & YFCC15M & 29.27 & 0.28 \\
OpenCLIP \cite{mehdi_cherti_reproducible_2023}  & ViT-B-16 \cite{ilharco_gabriel_2021_5143773} & MetaCLIP & 18.74 & 0.63 \\
OpenCLIP \cite{mehdi_cherti_reproducible_2023}  & ViT-B-16-plus-240 \cite{ilharco_gabriel_2021_5143773} & LAION400M & 13.45 & 0.72 \\
OpenCLIP \cite{mehdi_cherti_reproducible_2023}  & ViT-B-16-quickgelu \cite{ilharco_gabriel_2021_5143773} & DFN2B & 34.42 & 0.46 \\
OpenCLIP \cite{mehdi_cherti_reproducible_2023}  & xlm-roberta-base-VIT-B-32 \cite{ilharco_gabriel_2021_5143773} & LAION5B & 16.92 & 0.66 \\
OpenCLIP \cite{mehdi_cherti_reproducible_2023}  & ViT-B-32 & CommonPool & 144.54 & 0.08 \\
OpenCLIP \cite{mehdi_cherti_reproducible_2023}  & ViT-bigG-14 \cite{ilharco_gabriel_2021_5143773} & LAION2B & 8.18 & 0.81 \\
OpenCLIP \cite{mehdi_cherti_reproducible_2023}  & ViT-bigG-14-quickgelu \cite{ilharco_gabriel_2021_5143773} & MetaCLIP & \textbf{6.31} & 0.85 \\
OpenCLIP \cite{mehdi_cherti_reproducible_2023}  & ViT-g-14 \cite{ilharco_gabriel_2021_5143773} & LAION2B & 8.62 & 0.80 \\
OpenCLIP \cite{mehdi_cherti_reproducible_2023}  & convnext-xxlarge \cite{ilharco_gabriel_2021_5143773} & LAION2B AugReg & 6.71 & 0.83 \\
\hline
CLIPA \cite{li2024inverse} & ViT-H-14-CLIPA-336  & DataComp-1B & 9.25 & 0.82 \\
\hline
SigLIP \cite{zhai2023sigmoid} & ViT-L-16-SigLIP-384 \cite{ilharco_gabriel_2021_5143773} & WebLi & 12.93 & 0.71 \\
SigLIP \cite{zhai2023sigmoid} & ViT-SO400M-14-SigLIP-384 \cite{ilharco_gabriel_2021_5143773} & WebLi & 7.04 & 0.84 \\
SigLIP \cite{zhai2023sigmoid} & nillb-clip-large-siglip \cite{ilharco_gabriel_2021_5143773} & V1 & 96.46 & 0.32 \\
\hline

ViT-Lens \cite{weixian_lei_vit-lens_2023} & ViT-Lens-L & Mixed Sources & \textbf{7.78} & \textbf{0.85} \\ 
\hline
\end{tabular}
}
\end{table}

\textbf{Prompt Sensitivity Analysis:}
\label{app:subsec_prompt_sensiticity}
Table~\ref{tab:prompt_sensitivity} shows how different linguistic formulations affect time awareness performance across CLIP and EVA-CLIP models. The analysis demonstrates that prompt formulation is crucial for temporal prediction, with minimalistic prompts (P1, P2) performing significantly worse than more descriptive formulations.

\begin{table}[h!]
\caption{Prompt Sensitivity for CLIP and EVA-CLIP VLMs}
\label{tab:prompt_sensitivity}
\begin{center}
\footnotesize
\resizebox{\columnwidth}{!}{%
\begin{tabular}{l|cc|cc}
\hline
\textbf{Sentence Base} & \multicolumn{2}{c|}{\textbf{CLIP}} & \multicolumn{2}{c}{\textbf{EVA-CLIP}} \\
& \textbf{$\downarrow$MAE} & \textbf{$\uparrow$TAI} & \textbf{$\downarrow$MAE} & \textbf{$\uparrow$TAI} \\
\hline
P1 - [year] & 48.28 & 0.70 & 48.14 & 0.73 \\
P2 - Year [year] & 19.89 & 0.82 & 16.34 & 0.84 \\
P3 - Was released in the year [year] & 17.86 & 0.83 & 11.30 & 0.88 \\
P4 - Was invented in the year [year] & 19.34 & 0.81 & 9.18 & 0.89 \\
P5 - Was first introduced in the year [year] & 15.84 & 0.82 & 7.70 & 0.89 \\
P6 - Was created in the year [year] & 19.06 & 0.77 & 10.30 & 0.88 \\
P7 - Was built in the year [year] & \textbf{8.79} & \textbf{0.86} & \textbf{7.44} & \textbf{0.89} \\
P8 - First appeared in the year [year] & 15.19 & 0.82 & 14.24 & 0.85 \\
P9 - Emerged in the year [year] & 17.87 & 0.80 & 14.81 & 0.86 \\
\hline
\end{tabular}
}
\end{center}
\end{table}

\textbf{Class-wise performance:}
To examine how temporal awareness varies across object categories, we analyse the performance of EVA02-CLIP-L-14-336, one of our best-performing models, across the six classes in TIME10k (Table~\ref{tab:class_performance}). Cars and mobile phones, which have been photographed more often in recent periods, achieve excellent performance (MAE $\leq$ 2.65, TAI $\geq$ 0.95), while categories like music instruments and weapons, whose depictions often predate the era of photography and have less distinct times of first appearance cannot be positioned that accurately in time (MAE > 30, TAI < 0.5). These results reflect the natural evolution of visual documentation and the fact that the classes have different annual ranges (see last row of Table~\ref{tab:class_performance}). Notably, these class-specific scores do not directly average to those in Table~\ref{tab:vlms} due to differences in class cardinalities.

\begin{table}[]

\centering
\caption{Class-specific awareness to time.}
\label{tab:class_performance}
\renewcommand{\arraystretch}{1.2} 
\resizebox{\columnwidth}{!}{%

\begin{tabular}{l|cccccc}

\hline 
\textbf{Metric} & \textbf{Aircraft} & \textbf{Cars} & \textbf{Mobile} & \textbf{Music} & \textbf{Ships} & \textbf{Weapons} \\
& & & \textbf{Phones} & \textbf{Instruments} & & \textbf{\& Ammo} \\
\hline 
MAE $\downarrow$ & 14.25 & 2.64 & 2.65 & 32.94 & 18.12 & 33.50 \\
TAI $\uparrow$ & 0.76 & 0.95 & 0.95 & 0.24 & 0.76 & 0.49 \\
Annual range & [1893,2017] & [1888,2024] & [1984,2024] & [1715,2009] & [1744,1999] & [1939,2003] \\
\hline 
\end{tabular}
}
\vspace{-8pt}
\end{table}

\begin{figure}[H]
    \centering
    \includegraphics[width=\columnwidth]{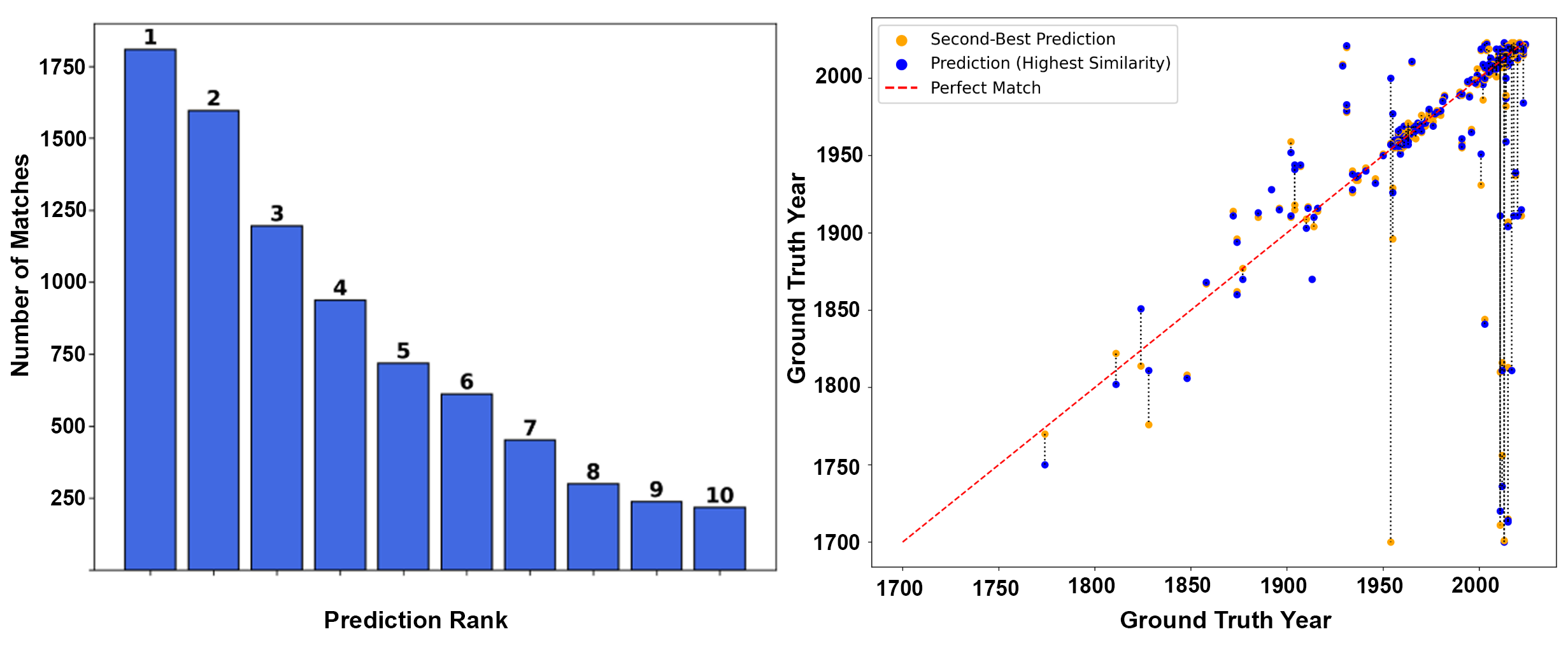}
    \caption{Left: The highest score (rank 1) is most often the correct year, followed by the second highest year (rank 2) etc. 
    Right: Ground truth (red diagonal) vs. predicted year for highest  (blue) and second-highest (orange) score. Vertical dotted lines represent their difference, which are mostly small.
    }
    \label{fig:ranking_scatter}
\end{figure}

\textbf{Validity of time probing:}
Time probing selects the year with the highest dot-product similarity among all candidates (see Equation~\ref{eq:probe_pred}). To assess the validity of this choice, we analyse the distribution of scores across probed years using two approaches with one of the best-performing model EVA02-CLIP-L-14-336. \textit{First}, we examine how often the top-ranked predictions match the ground truth year (Figure~\ref{fig:ranking_scatter}, left). The highest-scoring year (rank 1) is most frequently correct, followed by rank 2 and so on, confirming that the score reflects the probability of a correct assignment well. \textit{Second}, we analyse the relationship between the two top-scoring years and the ground truth (Figure~\ref{fig:ranking_scatter}, right). Both highest (blue) and second-highest (orange) scores align well with the true year (red diagonal). Some outliers exist, particularly for recent years, where either the first or second-highest prediction is incorrect. Overall, the score distribution suggests that time probing reliably identifies the most probable year of first appearance.

\subsection{Embedding Space Analysis}
\label{subsec:results_embeddinganalysis}

\begin{figure}[h]
    \centering
    \includegraphics[width=0.75\columnwidth]{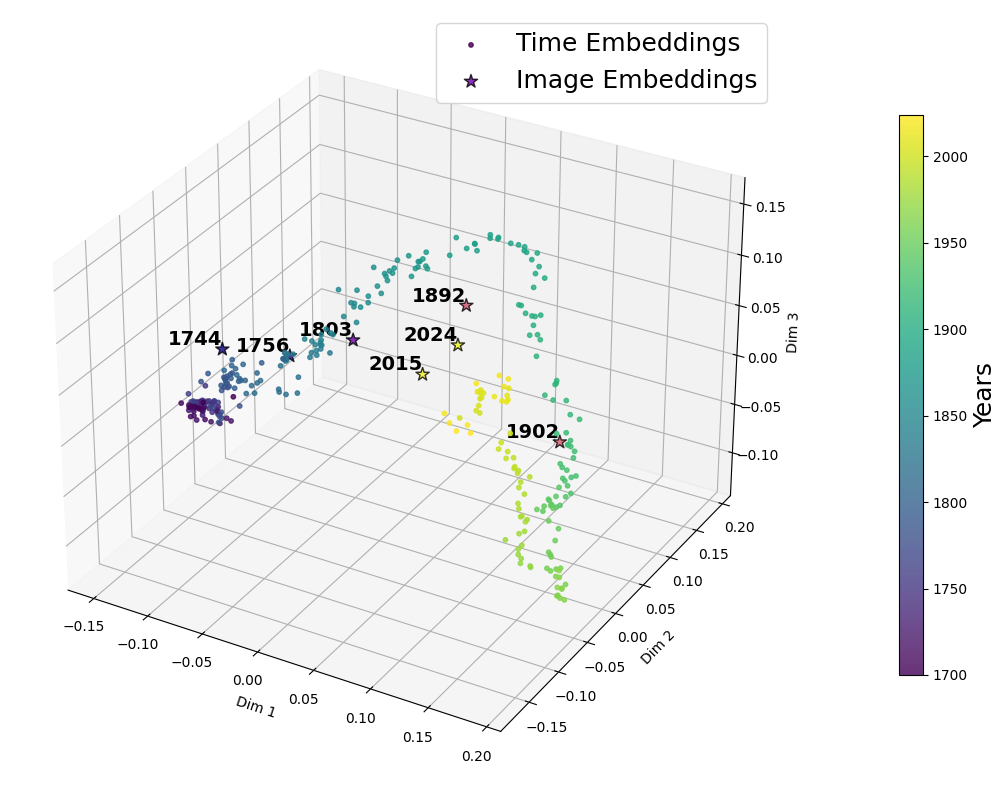}
    \caption{3D KPCA projection of CLIP time and image embeddings.}
    \label{fig:embedding_space}
\end{figure}

For embedding space analysis, we employ KPCA and UMAP and generate 1D, 2D and 3D projections of the time embeddings $\mathbf{T}_{1700}$,..., $\mathbf{T}_{2024}$ from the two selected representative models, ViT-B/32 (CLIP) and EVA02-CLIP-L-14-336 (EVA-CLIP). Figure~\ref{fig:embedding_space} shows the example of a 3D projection of time embeddings from the ViT-B/32 (CLIP) model obtained using KPCA. As hypothesised, the projections align along a lower-dimensional (non-linear) manifold-like structure. Colour encodes time from 1700 (violet) to 2024 (yellow), revealing a chronological progression of time along this structure. Thus, RQ2 can be answered positively in this case. We observe a similar behavior for EVA-CLIP, indicating that this structure is consistent across different VLM architectures. Notably, this contrasts with recent findings in language models where temporal information seems to be encoded in linear subspaces~\cite{gurnee2023language, heinzerling2024monotonic}, highlighting that VLMs organise temporal information differently, namely in low-dimensional non-linear structures.

To quantify chronological alignment, we generate 1D projections of $\mathbf{T}_{1700}$,..., $\mathbf{T}_{2024}$ using KPCA and UMAP with default parameters. Table~\ref{tab:ranking_metrics} presents the respective ranking scores after reducing the embeddings to 1D and comparing the resulting order with the ground truth. 
The ranking metrics indicate that the temporal order is preserved to a large degree (e.g. Spearman correlation $\rho$ of 0.96 for KPCA), showcasing a strong potential of the (projected) time embeddings to serve as a basis for deriving a timeline from it. 

\begin{table}[]
\footnotesize
\centering
\caption{Degree of chronological progression in 1D.}
\label{tab:ranking_metrics}
\begin{tabular}{l|cc|cc}
\hline
\textbf{Metric} & \multicolumn{2}{c|}{\textbf{CLIP}} & \multicolumn{2}{c}{\textbf{EVA-CLIP}} \\
                & \textbf{KPCA} & \textbf{UMAP} & \textbf{KPCA} & \textbf{UMAP} \\ \hline
$\rho$          & 0.96         & 0.80         & 0.92            & -0.70            \\
$\tau$          & 0.84         & 0.55         & 0.77            & -0.48            \\
$\delta_{\text{MNDL}}$ & 0.84        & 0.55         & 0.77            & 0.74            \\ \hline
\end{tabular}
\end{table}

Finally, we project randomly selected image embeddings into the 3D space from Figure~\ref{fig:embedding_space} (plotted as black stars with labels indicating their year). The image embeddings align well with the temporal structure, indicating temporal coherence across image and text modalities (RQ2).

\subsection{Time Modeling and Prediction}
\label{subsec:result_timeline}

\begin{figure}[]
    \centering
    \includegraphics[width=\columnwidth]{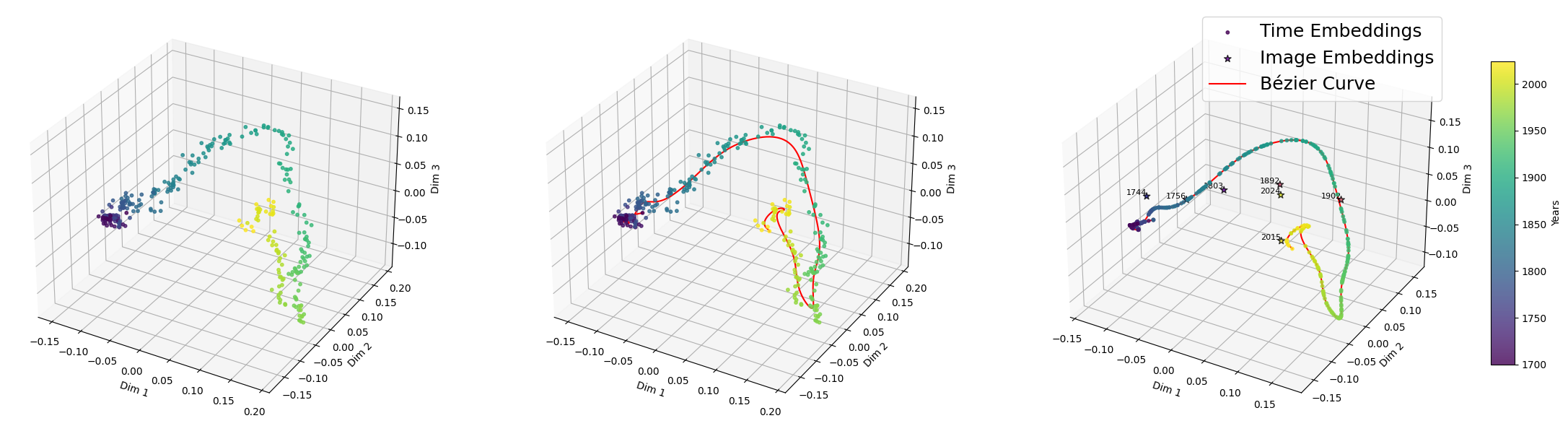} 
    \caption{Fitting a Bézier curve to the time embeddings from CLIP in a 3D projection space. (a) Time embeddings coloured by year with selected control points marked with diamonds. (b) Fitting the Bézier curve. (c) Mapping points to the Bézier curve to obtain a 1D timeline representation.}
    \label{fig:bezier_steps}
\end{figure}

We evaluate the two approaches for timeline modeling on ViT-B/32 (CLIP) and EVA02-CLIP-B/16 (EVA-CLIP). Figure~\ref{fig:bezier_steps} illustrates Bézier curve fitting in 3D projection space. Using \( \mathcal{K} = 200 \) chronologically sorted control points, we fit a 1D curve and map all time embeddings onto it to obtain a monotonic timeline. Figure~\ref{fig:bezier_steps}(c) further shows image embeddings projected into the same space (black stars with year label), showing that the image embeddings align well with the curve.

\begin{figure}[h]
    \centering
    \includegraphics[trim={0 10pt 0 22pt},clip,width=0.6\columnwidth]{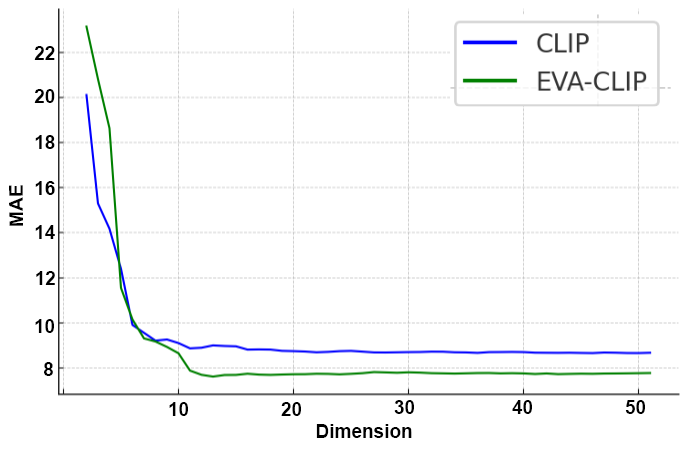}
    \caption{MAE per dimension for (EVA-)CLIP using KPCA.}
    \label{fig:mae_per_dim}
\end{figure}

\textbf{Projection space dimension.} For the Bézier approach the question arises, which projection space dimension is optimal, i.e. what is the minimal dimension necessary to capture the observed manifold-like structure. To answer this, we evaluate the Bézier approach for the entire TIME10k dataset for all possible projection dimensions and plot the MAE, see Figure~\ref{fig:mae_per_dim} for results on two different VLMs. With increasing dimension, MAE reduces rapidly and then stays more or less constant. According to Figure~\ref{fig:mae_per_dim},  for both VLMs around 13 dimensions are already sufficient to capture most of the temporal information. Higher dimensions do not yield significant benefits. This finding is particularly interesting as it suggests that temporal information in VLMs is encoded in a remarkably compact subspace compared to the original embedding dimension (typically $ \geq  512$). 


\textbf{Bezier approach variants.} We evaluate four variants of the proposed Bézier-based approach for temporal inference that (i) vary in embedding space dimension and (ii) in the temporal inference method. Bézier($\mathbf{R}^{N}$,NN) and Bézier($\mathbf{R}^{N}$,Int) operate in the original $N$-dimensional embedding space and use either nearest neighbor (``NN'') or linear interpolation (``Int'') to assign the closest year, for potentially finer granularity. Bézier($\mathbf{R}^{S}$,NN) and Bézier($\mathbf{R}^{S}$,Int) operate on an  $S$-dimensional subspace obtained via  KPCA. Based on the result in Figure~\ref{fig:mae_per_dim}, we set $S=13$.

\textbf{Chronological structure.} Table~\ref{tab:temporal_performance_comparison} compares time probing, UMAP, and the four Bézier-based methods for time prediction. Ranking scores (\(\rho, \tau, \delta_{\text{MNDL}}\)) between $0.96$ and $0.99$ confirm almost perfect chronological time representations. Note that for UMAP and EVA-CLIP the estimated timeline is inverted (negative correlation).

\textbf{Performance comparison.} The accuracy metrics (MAE, TAI) in Table~\ref{tab:temporal_performance_comparison} show that for CLIP the Bézier approaches clearly outperform UMAP and Time probing. 
For EVA-CLIP  the Bézier approaches in average outperform UMAP and time probing in terms of TAI. For MAE the Bézier approach (with dimensionality reduction) yields a similar performance level to time probing. Interpolation does not significantly affect overall accuracy compared to nearest neighbor matching. Dimensionality reduction by KPCA  consistently improves the Bézier approach, indicating that reducing dimensionality also removes unwanted factors or noise.

\begin{table}[ht]
\centering

\caption{Time prediction results for  UMAP- and Bézier-based timeline extraction compared to time probing (baseline). Results outperforming time probing are highlighted in bold.}
\label{tab:temporal_performance_comparison}
\resizebox{\columnwidth}{!}{
\begin{tabular}{l|cccccc}
\toprule
Metric & Time Probing & UMAP & Bézier($\mathbf{R}^{n}$,NN) & Bézier($\mathbf{R}^{n}$,Int) & Bézier($\mathbf{R}^{s}$,NN)  & Bézier($\mathbf{R}^{s}$,Int) \\
\midrule
\multicolumn{7}{c}{CLIP} \\
\midrule
MAE $\downarrow$ & 9.53 & 11.51 & \textbf{9.10} & \textbf{9.16} & \textbf{9.00} & \textbf{8.81} \\
TAI $\uparrow$ & 0.70 & \textbf{0.81} & \textbf{0.76} & \textbf{0.76} & \textbf{0.77} & \textbf{0.79} \\
Inference Time (ms) $\downarrow$ & \textbf{5} & 539 & 26 & 17 & 11 & 11 \\
\midrule
$\rho$  & -- & 0.99 & 0.99 & 0.99 & 0.99 & 0.99 \\
$\tau$  & -- & 0.98 & 0.99 & 0.99 & 0.99 & 0.99 \\
$\delta_{\text{MNDL}}$  & -- & 0.98 & 0.99 & 0.99 & 0.99 & 0.99 \\
\midrule
\multicolumn{7}{c}{EVA-CLIP} \\
\midrule
MAE $\downarrow$ & \textbf{7.67} & 15.49 & 9.58 & 9.61 & {7.76} & {7.77} \\
TAI $\uparrow$ & 0.82 & 0.74 & \textbf{0.85} & 0.72 & \textbf{0.84} & \textbf{0.84} \\
Inference Time (ms) $\downarrow$ & 10 & 554 & \textbf{9} & \textbf{9} & 12 & 11 \\
\midrule
$\rho$  & -- & -0.99 & 0.99 & 0.99 & 0.99 & 0.99 \\
$\tau$  & -- & -0.96 & 0.98 & 0.98 & 0.98 & 0.98 \\
$\delta_{\text{MNDL}}$  & -- & -0.96 & 0.98 & 0.98 & 0.98 & 0.98 \\
\bottomrule
\end{tabular}
}
\end{table}

\textbf{Runtime comparison.} Experiments were conducted on an NVIDIA Tesla V100 GPU (32GB VRAM) with an AMD EPYC 9334 32-core processor and 64GB system memory. Inference times differ substantially between methods (see Table~\ref{tab:temporal_performance_comparison}). The UMAP approach requires significantly more time (539-554ms per prediction) due to the projection process, while Bézier-based approaches achieve superior efficiency at 9-26ms and thus reach a competitive speed to time probing  with the added benefit of getting an explicit timeline representation that can be efficiently re-used for time reasoning.

\section{Conclusion}

We presented a first systematic investigation of temporal awareness in vision-language models. We systematically evaluated 37 state-of-the-art VLMs with a new benchmark dataset (TIME10k) and found that modern models possess substantial temporal awareness, although performance varies significantly with training data quality and backbone model complexity. Most remarkably, we discovered that temporal information is encoded in a compact, low-dimensional manifold with strong chronological structure within the high-dimensional embedding space and developed novel approaches for the extraction of an explicit timeline representation based on UMAP and Beziér curve approximation. 
Still, there are certain limitations of our investigation motivating future work: (i) the effect of class imbalances and annual ranges needs to be further investigated; (ii) TIME10k  in future should be extended with classes representing human-centric aspects (e.g. clothing); (iii) the evaluation of timeline extraction methods should be extended to more VLMs as well as generative VLMs to see if they exhibit similar temporal awareness (which we assume as they base on similar encoders). Ultimately, our work raises a fundamental question: if VLMs can implicitly capture and structure temporal information so effectively, can this approach generalize to other ordinal problems? 

\section*{Acknowledgements}
This work was funded by the Austrian Science Fund (FWF) under the Visual Heritage PhD Program doctoral grant [10.55776/DFH37].

\printbibliography

\end{document}